\documentclass[letterpaper]{article} 
\usepackage{aaai2026}  
\usepackage{times}  
\usepackage{helvet}  
\usepackage{courier}  
\usepackage[hyphens]{url}  
\usepackage{graphicx} 
\usepackage{natbib}  
\usepackage{caption} 
\usepackage{algorithm}
\usepackage{algorithmic}

\usepackage{multirow}
\usepackage{booktabs} 
\usepackage{array}    
\newcommand{\corrauthor}{\thanks{Corresponding authors.}} 
\newcommand{\corrauthormark}{\footnotemark[\value{footnote}]} 

\usepackage{cuted}      
\usepackage{caption}    
\usepackage{amsmath}

\usepackage{newfloat}
\usepackage{listings}
\DeclareCaptionStyle{ruled}{labelfont=normalfont,labelsep=colon,strut=off} 
\floatstyle{ruled}
\newfloat{listing}{tb}{lst}{}
\floatname{listing}{Listing}
\title{SatireDecoder: Visual Cascaded Decoupling for Enhancing\\Satirical Image Comprehension}

\author{
    Yue Jiang\textsuperscript{\rm 1}\equalcontrib,
    Haiwei Xue\textsuperscript{\rm 2, \rm 5, \rm 6}\equalcontrib,
    Minghao Han\textsuperscript{\rm 1},
    Mingcheng Li\textsuperscript{\rm 1},
    Xiaolu Hou\textsuperscript{\rm 1},\\
    Dingkang Yang\textsuperscript{\rm 1,\rm 3}\corrauthor, 
    Lihua Zhang\textsuperscript{\rm 1}\corrauthormark,    
    Xu Zheng\textsuperscript{\rm 4, \rm 5, \rm 6}\corrauthormark        
}

\affiliations {
    \textsuperscript{\rm 1}College of Intelligent Robotics and Advanced Manufacturing, Fudan University\\
    \textsuperscript{\rm 2}Tsinghua University\\ 
    \textsuperscript{\rm 3}ByteDance\\
    \textsuperscript{\rm 4}INSAIT, Sofia University “St. Kliment Ohridski”\\
    \textsuperscript{\rm 5}The Hong Kong University of Science and Technology\\
    \textsuperscript{\rm 6}The Hong Kong University of Science and Technology (Guangzhou)\\
    jiangyue23@m.fudan.edu.cn,
    winniygd@outlook.com,\\
    \{lihuazhang, dkyang20\}@fudan.edu.cn,
    zhengxu128@gmail.com
}

\begin{document}
\maketitle



\begin{abstract}
Satire, a form of artistic expression combining humor with implicit critique, holds significant social value by illuminating societal issues. Despite its cultural and societal significance, satire comprehension, particularly in purely visual forms, remains a challenging task for current vision-language models. This task requires not only detecting satire but also deciphering its nuanced meaning and identifying the implicated entities. Existing models often fail to effectively integrate local entity relationships with global context, leading to misinterpretation, comprehension biases, and hallucinations. To address these limitations, we propose SatireDecoder, a training-free framework designed to enhance satirical image comprehension. Our approach proposes a multi-agent system performing visual cascaded decoupling to decompose images into fine-grained local and global semantic representations. In addition, we introduce a chain-of-thought reasoning strategy guided by uncertainty analysis, which breaks down the complex satire comprehension process into sequential subtasks with minimized uncertainty. Our method significantly improves interpretive accuracy while reducing hallucinations. Experimental results validate that SatireDecoder outperforms existing baselines in comprehending visual satire, offering a promising direction for vision-language reasoning in nuanced, high-level semantic tasks.
\end{abstract}

\begin{figure*}[t]
  \centering
  \includegraphics[width=\textwidth]{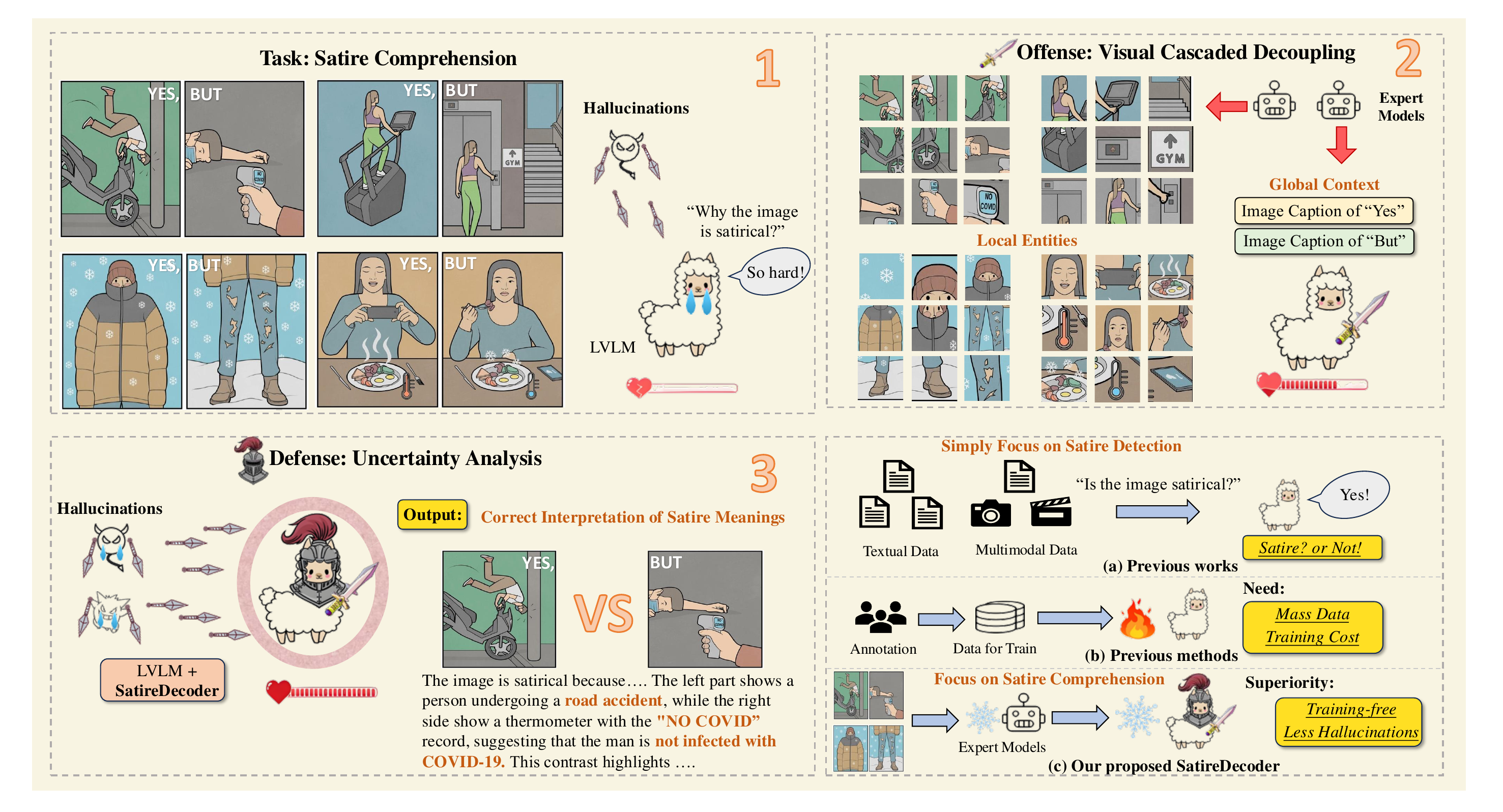}
  \caption{SatireDecoder consists of a multi-agent visual cascaded decoupling module and an uncertainty analysis strategy, which together help MLLMs capture discrepancies between local entities and the global context, thereby enhancing models' ability to comprehend satirical images.}
  \label{fig-teaser}
\end{figure*}

\section{Introduction}
Satirical images often rely on deliberately contradictory or conflicting scenes to convey the deep semantics, blending humor with subtle critique~\cite{del2017automatic}. People frequently employ satirical images on social media platforms to express their attitudes toward social phenomena or trending events. Consequently, comprehending the satirical semantics inherent in images holds considerable scholarly significance. Understanding satirical images requires identifying inherent conflicts and analyzing the interaction between local entities and global contexts to infer the deep semantics.

As illustrated in Figure~\ref{fig-teaser}, previous works~\cite{yu2023mmoe, zhu2024tfcd, xie2024moba, yue2024sarcnet} have exclusively focused on simplistic satire detection, which is an easy binary classification, neglecting the more challenging task of understanding and interpreting the deep satirical semantics inherent in images. Due to the difference between the image's satirical and surface meanings, comprehending its deep semantics necessitates a thorough analysis of the relationships between local entities and the global context to uncover contradictory or incongruent components.
Moreover, previous methods~\cite{wu2018thu_ngn, chen2024can, chen2024efficiency} has depended on large datasets and high training costs, thereby suffering from substantial overhead and limited portability. 
In addition, despite the remarkable success of current multimodal large language models (MLLMs) in multimodal tasks, several popular MLLMs exhibit significant limitations in comprehending image deep semantics beyond the surface meanings~\cite{tu-etal-2024-multiple, chang2024nykmswellannotatedmultimodalmetaphor}. 
MLLMs tend to overlook or fabricate local entities and crucial details within images~\cite{chen2024miss, chen2024detecting, yang2024towards, yang2024asynchronous}, resulting in hallucination issues and the misinterpretation of satirical semantics~\cite{leng2023mitigatingobjecthallucinationslarge}.
Furthermore, MLLMs lack a step-by-step inference process from local entities to global context during image understanding, creating significant challenges in grasping the relationships between visual elements and the deep semantics of satire~\cite{huang2025survey}.

To address the limitations of MLLMs in comprehending satirical images, we propose SatireDecoder, a novel training-free framework illustrated in Figure~\ref{fig-teaser}. SatireDecoder employs a multi-agent visual cascaded decoupling mechanism to decompose images into fine-grained semantic representations, effectively capturing both local entity features and global contextual cues. This design further enables the identification of semantic discrepancies, contradictions, and incongruities—key characteristics of visual satire. Subsequently, a Chain-of-Thought (CoT) reasoning strategy guided by uncertainty analysis decomposes the complex satire comprehension process into sequential subtasks with minimized uncertainty, improving interpretability while mitigating hallucinations.
SatireDecoder can be seamlessly integrated into various MLLM baselines and consistently enhances their performance in satire understanding. Extensive experiments and ablation studies further validate the effectiveness of SatireDecoder and the contribution of each component within the framework. Our main contributions are summarized as follows:
\begin{itemize}
    \item We propose SatireDecoder, a novel training-free framework that leverages multi-agent collaboration for visual cascaded decoupling, decomposing complex satirical images into fine-grained semantic representations across local and global levels, enhancing visual perception.
\end{itemize}

\begin{itemize}
    \item We design a CoT reasoning, guiding MLLMs to decompose the satire comprehension into three subtasks: identifying local entities, understanding global context, and inferring satirical intent. This approach improves interpretability and demonstrates generalization ability in complex, resource-constrained visual reasoning tasks.
    
\end{itemize}

\begin{itemize}
    \item We introduce an uncertainty-guided inference method, which quantifies the discrepancy between the outputs of MLLMs and multi-agents for shared subtasks. By minimizing the uncertainty score, our method reduces hallucinations and enhances the robustness of final satire interpretation, offering a novel perspective on interpretable model optimization through uncertainty analysis.
    
\end{itemize}

\section{Related Work}

\subsection{Text-based Satire Analysis}
Satire comprehension is crucial for sentiment analysis and identifying harmful comments. 
Previous studies on satire in text modality focus on the satire detection~\cite{joshi2017automatic}. Several supervised approaches are applied to the tasks, including traditional machine learning methods with lexical features~\cite{ptaek2014sarcasm, bouazizi2015sarcasm} and deep learning method~\cite{wu2018thu_ngn}. 
Moreover, various emotional~\cite{thu2018implementation, li2024toward}, psychological~\cite{del2017automatic}, and linguistic features~\cite{yang2017satirical} are incorporated to enhance satire detection. The effect of contextual inconsistencies in satire detection has also been explored~\cite{joshi2015harnessing}, emphasizing the importance of semantic and pragmatic factors.
However, images have emerged as a pivotal medium for information dissemination on social media. Text-based satire analysis is inadequate for fully capturing the satirical content present in contemporary media.

\subsection{Multimodal Satire Analysis}
Multimodal satire presents unique challenges due to the interplay between modalities. Previous research has focused primarily on detection and classification tasks~\cite{cai2019multi, castro2019towards, yu2023mmoe, zhu2024tfcd, xie2024moba, Xue_2025}. Early approaches~\cite{schifanella2016detecting, das2018sarcasm} employ traditional computer vision techniques combined with text analysis for meme classification. Recent methods have evolved toward deep learning architectures, incorporating pre-trained vision and language models for enhanced feature extraction~\cite{cai2019multi, bharti2022multimodal}. However, these approaches typically treat satire detection as a binary classification task, without addressing the deeper understanding of satirical elements and their interactions. Satire comprehension involves identifying contextual inconsistencies and reasoning about the satirical intent, evaluated by how well the visual differences and motivations are captured.

In the general domain, numerous low-cost, training-free methods exist to enhance the multimodal comprehension capabilities of MLLMs. Several works utilize multi-agent systems~\cite{li2025mccd, jiang2024multi} or incorporate specialized external models~\cite{zeng2022socraticmodelscomposingzeroshot, hyun2024smilemultimodaldatasetunderstanding} to augment multimodal processing. Some approaches~\cite{jiang2024comt, bi2025cot, bi2025llava, zhao2025can} further integrate chain-of-thought reasoning to guide the model in decomposing understanding into sequential steps. Regarding the mitigation of hallucinations in MLLMs, mainstream methods~\cite{leng2024vcd,zhang2025diffusionvcd,wang2025ascd, wang2024conu} involve optimizing the model during decoding by applying principles of contrastive learning, among others.

\begin{figure}[t] 
    \centering
    \includegraphics[width=0.7\linewidth]{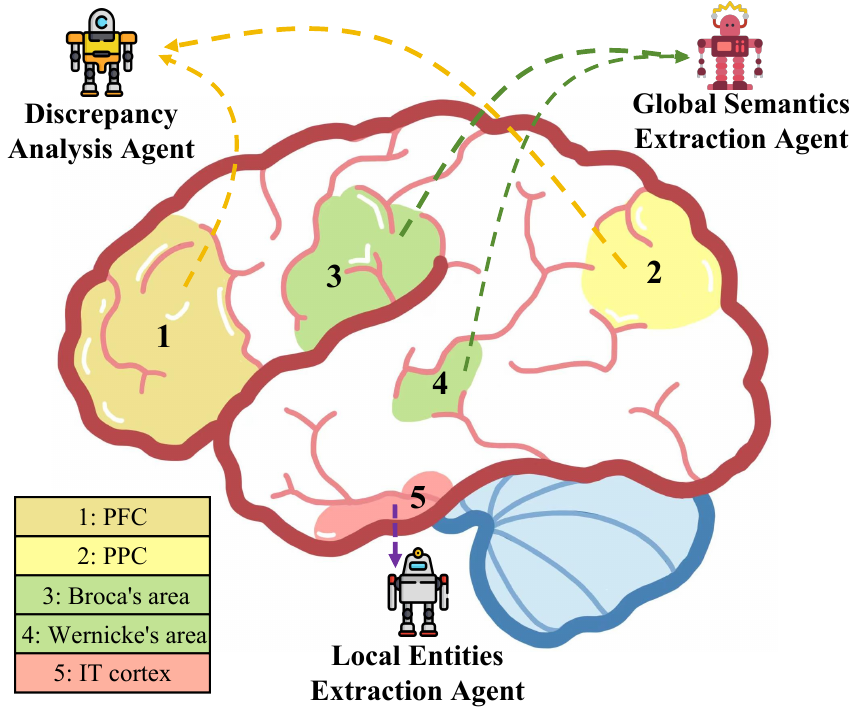} 
    \caption{In the multi-agent system, the Local Entities Extraction Agent simulates the InferoTemporal cortex (IT cortex)~\cite{grill2004human}, while the Global Semantics Extraction Agent simulates the Broca's area and the Wernicke's area~\cite{jancke2021language}. Additionally, the Discrepancy Analysis Agent imitates the function of the PreFrontal Cortex (PFC) and the Posterior Parietal Cortex (PPC)~\cite{grill2004human}.}

    \label{fig-2}
\end{figure}


\begin{figure*}[t] 
    \centering
    \includegraphics[width=\textwidth]{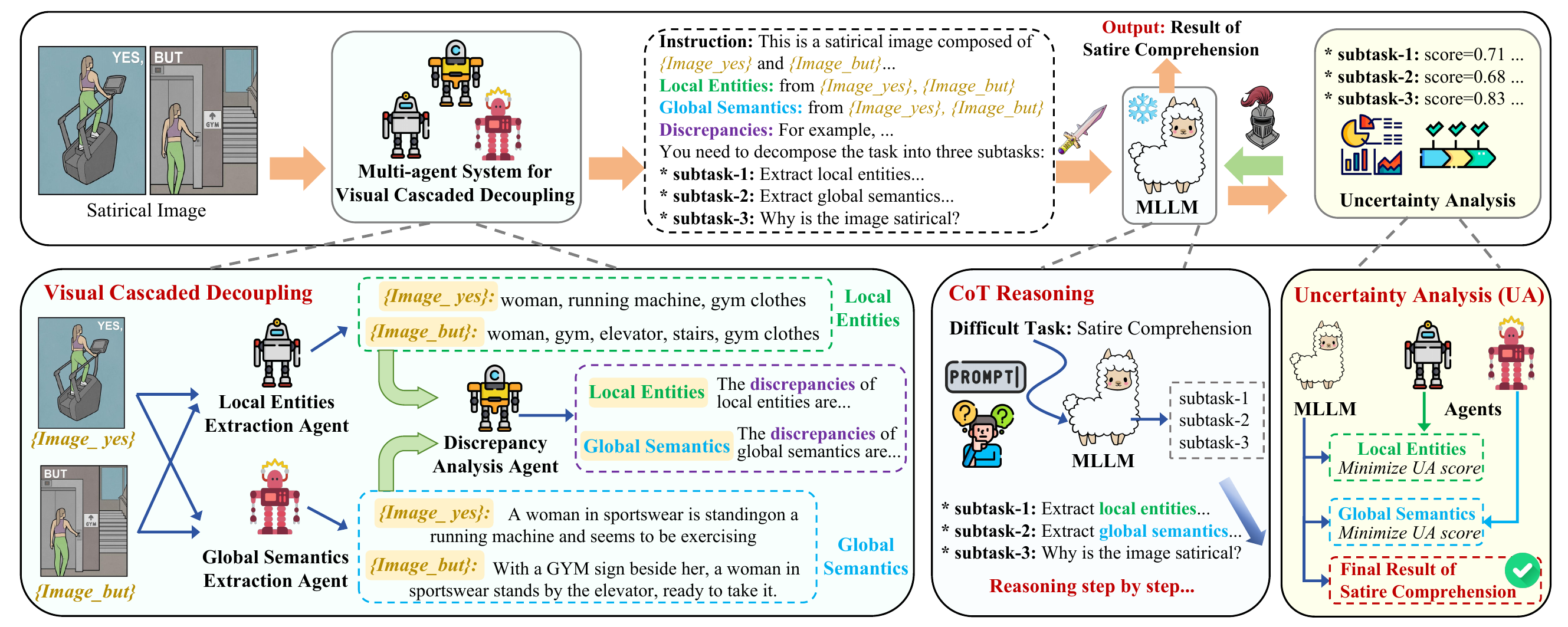} 
    \caption{
\textbf{Framework of SatireDecoder.} (1) Visual Cascaded Decoupling: A multi-agent system decouples the satirical image into fine-grained semantic representations. (2) Prompt Construction and CoT Reasoning: The fine-grained semantic representations form a structured prompt for CoT reasoning, breaking complex satire comprehension into three subtasks. (3) Inference Optimization with Uncertainty Analysis: progressively reduces uncertainty and hallucinations during reasoning.
}

    \label{fig-overview}
\end{figure*}

\subsection{Datasets for Satire Analysis}
Beyond conventional text-based satire detection, several new datasets have been developed for multimodal scenes.
MMSD~\cite{cai2019multi}, MMSD2.0~\cite{qin2023mmsd2}, MUStARD~\cite{castro2019towards}, and SarcNet~\cite{yue2024sarcnet} focus solely on satire detection, failing to assess the ability of MLLMs to comprehend the deep semantics of satire. NYK-MS~\cite{chang2024nykmswellannotatedmultimodalmetaphor} consists of more than 1,500 cartoon-caption pairs, supporting both satire detection and comprehension. But in NYK-MS, satire is conveyed through the combination of image and textual descriptions, which simplifies the comprehension task.
In our paper, we utilize YesBut~\cite{nandy2024yesbuthighqualityannotatedmultimodal}, which is the only current dataset specifically designed for satire comprehension and without image captions. Each satirical image in YesBut is structured in a ``Yes, But'' format, where the left half of the image depicts a normal scene, juxtaposed with a conflicting scene on the right, which together constitute the deep semantics of satire. In YesBut, satire is conveyed solely through visual information, without accompanying any textual clues, posing a unique challenge for MLLMs.

\section{Methodology}
The framework of SatireDecoder is depicted in Figure~\ref{fig-overview}. 
Inspired by the human perception paradigm on visual information~\cite{yang2023target, mischler2024contextual} and the multi-agent collaboration~\cite{jiang2025danmakutppbench, li2025mccd}, we propose a multi-agent system for visual cascaded decoupling to decompose the visual input into fine-grained representations. Based on the multi-agent collaboration, a CoT prompt is constructed, which is combined with the fine-grained information from the image. Catalyzed by the CoT prompt, MLLM decomposed the complex satire comprehension task into three subtasks, followed by the uncertainty analysis strategy to minimize the uncertainty score during the three-step inference, which efficiently mitigates hallucinations.

\subsection{Multi-agent-based Visual Cascaded Decoupling}
Related studies~\cite{mischler2024contextual, bullier2001integrated, grill2004human} have demonstrated that distinct regions of the cerebral cortex exhibit functional specificity in visual information processing. Inspired by this, we propose a biologically inspired multi-agent framework to simulate this mechanism. As shown in Figure~\ref{fig-2}, the agents are employed to play distinct roles, imitating different areas of the cerebral cortex. 
To extract local entity details, we employ the Local Entities Extraction Agent \(LE\) for image tagging, simulating IT cortex in the cerebral cortex, which includes object-selective regions and plays a crucial role in object recognition~\cite{bullier2001integrated, grill2004human}. The satirical image consists of two parts, \(\{Image\_yes\}\) and \(\{Image\_but\}\), denoted as \(I_y\) and \(I_b\) respectively. The \(LE\) is utilized to effectively detect and assign relevant tags to local entities within \(I_y\) and \(I_b\), and results are denoted as $LE_y = LE (I_y)$, and $LE_b = LE (I_b)$.

To grasp the global context, we leverage the Global Semantics Extraction Agent \(GS\) for image captioning, simulating PPC and PFC in the cerebral cortex. The two regions play essential roles in the integration of global visual information from complex scenes and the facilitation of high-level cognition and decision-making~\cite{bullier2001integrated, grill2004human}. The \(GS\) is utilized to process the two contrasting scenes \(I_y\) and \(I_b\), and the results can be represented as $GS_y = GS (I_y)$, and $GS_b = GS (I_b)$.

To contrast the subtle discrepancies and inconsistencies between the two scenes \(I_y\) and \(I_b\) depicted in the satirical image, we employed the Discrepancy Analysis Agent \(DA\) to simulate the Broca's and Wernicke's areas~\cite{jancke2021language} in the cerebral cortex, controlling complex vision-language comprehension~\cite{bullier2001integrated, grill2004human} and analyzing the outputs of \(LE\) and \(GS\) agents. The discrepancies of local entities \(D_l\) and global semantics \(D_g\) are represented as $D_l = DA(LE_y, LE_b)$ and $D_g = DA(GS_y, GS_b)$.

Within our multi-agent framework, to optimally balance cost and performance, we designate RAM~\cite{zhang2023recognizeanythingstrongimage} and BLIP~\cite{li2022blip} to play the role of Local Entities Extraction Agent and Global Semantics Extraction Agent, respectively. RAM is a specialized model for image tagging, engineered to accurately identify a wide range of common object categories within an input image, while BLIP is a pretraining model developed to bridge the gap between visual and linguistic understanding. It is capable of generating fluent and coherent descriptions that accurately reflect the content of an image. For the two elementary visual-semantic tasks, RAM and BLIP can match the performance of many MLLMs, paving the way for broader adoption of our training-free method. Concurrently, Qwen2~\cite{team2024qwen2} is employed as the Discrepancy Analysis Agent to undertake more complex, comprehension-intensive higher-level semantic tasks by harnessing the robust language understanding capabilities. By leveraging the multi-agent collaboration, we systematically decouple the satirical images into fine-grained semantic representations, as shown in Figure~\ref{fig-overview}, enhancing the perception of visual information from local and global perspectives.

\subsection{Prompt construction and CoT Reasoning}

The fine-grained semantic representations decoupled from images, including $\{LE_y$, $LE_b$, $GS_y$, $GS_b$, $D_l$, $D_g\}$, are employed to construct a CoT prompt to decompose the complex satire comprehension task into three subtasks, sequentially focusing on: local entity extraction, global semantic extraction, and satirical meaning inference, which facilitate step-by-step inference from local and global perspectives and guiding MLLM to focus on elementary details and explore the transformation or potential incongruous elements between the two scenes \(I_y\) and \(I_b\) when interpreting satirical images.
By further analyzing the incongruities within their social or cultural contexts, MLLM is induced to explore the deep semantics of satire in conjunction with social issues. Herein, we denote the results of the subtasks as \(R_1\), \(R_2\), and \(R_3\). The overview of prompt is depicted in Figure~\ref{fig-overview}.

\begin{table*}[t]
\setlength{\tabcolsep}{6pt}
\centering

\begin{tabular}{l|c|ccccc|ccccc}
\hline
\multicolumn{1}{c|}{\multirow{2}{*}{\textbf{Model}}} & \multicolumn{1}{c|}{\multirow{2}{*}{\textbf{Size}}} & \multicolumn{5}{c|}{\textbf{Automatic Evaluation}~$\uparrow$} & \multicolumn{5}{c}{\textbf{User Study}~$\uparrow$}         \\ \cline{3-12} 
\multicolumn{1}{c|}{}                       & \multicolumn{1}{c|}{}                      & BLEU   & R-L   & MT   & BERT  & AVE  & Correct & Length & Complete & Faithful & AVE \\
\hline
MiniGPT4    & 7B        & 0.002          & 0.143           & 0.156          & 0.828             & 0.282 &15.67 &3.00 &2.67 &19.33  & 10.18            \\
GPT4      & -           & 0.003          & 0.151           & 0.219          & 0.852             & 0.306 &58.00 &31.67 &37.00 &45.33    & 43.00            \\
Kosmos-2 & -          & 0.011          & 0.202           & 0.197          & 0.867             & 0.319  &15.33 &5.33 &7.67 & 9.00   & 9.33          \\
Gemini     & -     & 0.008          & 0.190           & 0.238          & 0.853             & 0.322     &46.67 &56.33 &52.00 &49.67 & 51.17           \\ \hline
LLaVA  & 7B           & 0.011          & 0.180           & 0.225          & 0.859             & 0.319    &25.67 &19.67 &23.00 &26.33  & 23.67           \\
LLaVA + $\clubsuit$ & 7B   & 0.034 & 0.239  & 0.270  & 0.869   & 0.353   &62.33 &21.33 &42.67 &59.67   &46.50           \\ 
LLaVA  & 13B           & 0.014          & 0.197           & 0.226          & 0.860             & 0.324    &28.33 &21.00 &27.67 &29.33  & 26.58          \\
LLaVA + $\clubsuit$ & 13B   & 0.037 & 0.240  & 0.273  & 0.870   & 0.355   &62.00 &20.67 &50.33 &56.00   &47.25           \\ 
LLaVA-Next  & 7B           & 0.013          & 0.189           & 0.230          & 0.861             & 0.323    &27.33 &21.00 &25.67 &27.67  &25.42           \\
LLaVA-Next + $\clubsuit$ & 7B   & 0.035 & \textbf{0.249}  & 0.276  & 0.872   & 0.356  &65.00 &20.67 &48.33 & 58.67    &48.17         \\

Qwen-VL & 7B           & 0.014         & 0.177           & 0.219          & 0.854             & 0.316   &31.00 &21.33 &29.67 &34.33   &29.08          \\
Qwen-VL + $\clubsuit$ & 7B   & 0.030 & 0.229  & 0.271  & 0.869   & 0.350   &56.00 &25.33 &47.67 &61.33    &47.58          \\

Qwen2.5-VL  & 7B           & 0.026          & 0.214           & 0.240          & 0.865             & 0.336    &61.33 &49.67 &52.00 &54.33   &54.33          \\ 
Qwen2.5-VL + $\clubsuit$ & 7B   & \textbf{0.038} & 0.247  & \textbf{0.279}  & \textbf{0.873}   & \textbf{0.360}  &\textbf{71.33} &\textbf{50.33} &\textbf{64.67} &\textbf{72.00} & \textbf{64.58}             \\
\hline
\end{tabular}

\caption{Comparison of different baselines in satirical image comprehension. We conduct a user study and automatic evaluation of NLG metrics. The symbol ``$\clubsuit$'' represents SatireDecoder. ``R-L'', ``MT'', ``BERT'', and ``AVE'' stand for ROUGE-L, METEOR, BERTScore, and the average scores computed across other metrics. User study is based on four criteria: correctness of the satire interpretation, appropriateness of the interpretation length, visual completeness interpretation, and faithfulness to the visual objects. Each criterion is evaluated by three users with a binary ``yes'' or ``no'' judgment. Baselines include MiniGPT4~\cite{minigpt4}, GPT4~\cite{gpt4}, Kosmos-2~\cite{kosmos2}, Gemini~\cite{gemini}, LLaVA~\cite{liu2024llava}, LLaVA-Next~\cite{liu2024llavanext}, Qwen-VL~\cite{Qwen-VL}, Qwen2.5-VL~\cite{Qwen2.5-VL}}

\label{tab-1}
\end{table*}

\subsection{Inference Optimization with Uncertainty Analysis}
In complex scenes, current MLLMs show a propensity to misinterpret crucial image elements, overlook local entities, and fabricate non-existent objects, leading to hallucination issues~\cite{ huang2025survey} and unreliable outputs. Uncertainty refers to the level of confidence or the degree of unpredictability associated with the outputs of models and has been proven to have a significant effect on hallucination issues~\cite{zhou2024analyzingmitigatingobjecthallucination}. To mitigate hallucinations while comprehending satirical images, we propose an uncertainty analysis strategy during inference.

After the visual cascaded decoupling of satirical images, the Chain-of-Thought prompt directs the MLLM to orderly perform three subtasks in the process of satire comprehension inference. During this process, uncertainty analysis scores (UA scores) are computed for the results \(R_1\) (about local entities) and \(R_2\) (about global semantics) generated by the subtask-1 and subtask-2, against the standardized outputs \(LE\_R_1\) and \(GS\_R_2\) of Local Entities Extraction Agent \(LE\) and Global Semantics Extraction Agent \(GS\), respectively. This procedure is repeated multiple times with varying model temperature settings of MLLM to minimize the UA scores of subtask-1 and subtask-2. Temperature is a parameter employed to regulate the creativity level~\cite{zhu2024hot} in text generation by language models. Given the logits \(Z_{i}\) for each candidate word, the corresponding probability distribution \(P(i)\) is computed as:
\begin{equation}
  P(i) = \frac{e^{z_i / Temp}}{\sum_{j} e^{z_j / Temp}}
  \label{eq:temperature}
\end{equation}
where \( Temp \) denotes model temperature, \(i\) represents the index of the target word under evaluation. and \(j\) corresponds to the index of all words in the vocabulary. Temperature modulates the probability distribution of the model, influencing the generation process by making the output content either more focused and deterministic or more random and diverse. Increasing temperature promotes greater diversity in generated content, revealing deeper comprehension and potentially hidden meanings. However, higher temperature also increases the risk of hallucinations and logical incoherence. Conversely, lower temperatures provide more stable and coherent outputs, reducing randomness but potentially overlooking subtle or latent implications.
Our method reduces the model's uncertainty regarding intermediate reasoning steps during complex reasoning tasks, thereby achieving the objective of controlling the reasoning path to obtain better responses for the final subtask-3 (about satire comprehension) and mitigate hallucinations inherent in multi-step inference processes.

Specifically, the results \(LE\_R_1\) and \(R_1\) derived from the Local Entities Extraction Agent and MLLM subtask-1 are sets of local entity tags within satirical images. The uncertainty is quantified as the opposite number of the Jaccard similarity coefficient~\cite{jaccard1912distribution}, as follows:

\begin{equation}
U_1 = min\{Temp(- \frac{|LE\_R_1 \cap R_1|}{|LE\_R_1 \cup R_1|})\}
\end{equation}

\begin{table*}[t]
\setlength{\tabcolsep}{6pt}
\centering
\begin{tabular}{l|ccccc|cc}
\hline 
\textbf{Model}            & \textbf{Correct}~$\uparrow$   & \textbf{Length}~$\uparrow$    & \textbf{Complete}~$\uparrow$  & \textbf{Faithful}~$\uparrow$ & \textbf{AVE}~$\uparrow$ & \textbf{CHAIR\_i}~$\downarrow$ & \textbf{CHAIR\_s}~$\downarrow$ \\ \hline
LLaVA+$\clubsuit$     &62.33 &21.33 &42.67 &59.67  & 46.50 & 36.53 & 41.02          \\
LLaVA+$\clubsuit$ (w/o UA)    &43.33 &20.00 &28.67 &47.33  & 34.83 & 55.39 & 59.17  \\

LLaVA-Next+$\clubsuit$    &65.00 &20.67 &48.33 & 58.67     & 48.17 & 34.80 & 39.75       \\ 
LLaVA-Next+$\clubsuit$ (w/o UA)      & 47.67    & 21.00     & 36.33     & 41.00 & 36.50 & 49.53 & 55.24 \\


Qwen-VL+$\clubsuit$   &56.00 &25.33 &47.67 &61.33  & 47.58 & 39.83 & 49.01       \\ 
Qwen-VL+$\clubsuit$ (w/o UA)      &34.67 &22.00 &29.67 &45.00 & 32.84 & 54.79 & 59.64 \\

Qwen2.5-VL+$\clubsuit$     &\textbf{71.33} &\textbf{50.33} &\textbf{64.67}&\textbf{72.00}       & \textbf{64.58} & \textbf{26.90} & \textbf{35.62}      \\ 
Qwen2.5-VL+$\clubsuit$ (w/o UA)   &65.67 &49.33 &54.00 &59.67 & 57.17 & 39.75 & 49.28 \\
\hline
\end{tabular}
\caption{Ablation study of the uncertainty analysis in SatireDecoder. To evaluate the effect of hallucination mitigation, the ablation experiment is based on the user study and CHAIR metrics from the object and sentence levels. The symbol ``$\clubsuit$'' represents SatireDecoder. ``w/o'' stands for ``without''. ``UA'' represents uncertainty analysis during inference.}
\label{tab-2}
\end{table*}

Furthermore, the results \(GS\_R_2\) and \(R_2\) derived from the Global Semantics Agent and MLLM subtask-2 are captions of satirical images. The uncertainty is quantified as the opposite number of the BERTScore~\cite{bertscore}, which leverages contextual embeddings from pre-trained language models~\cite{devlin2019bert, liu2019roberta} to measure the semantic similarity between a candidate text and a reference text. The UA score of \(GS\_R_2\) and \(R_2\) can be expressed as:

\begin{equation}
U_2 = min\{Temp(- BERTScore(GS\_R_2,~R_2))\}
\end{equation}

By controlling the model temperature hyperparameter to minimize uncertainty in the CoT reasoning, the result for subtask-3 exhibiting the least uncertainty is obtained as the final response for the satire comprehension task in our study.

\section{Experimental Setup}
\textbf{Baseline.} To ensure the consistency of the experiments, we follow~\cite{nandy2024yesbuthighqualityannotatedmultimodal}, utilizing the baselines including MiniGPT4~\cite{minigpt4}, GPT4~\cite{gpt4}, Kosmos-2~\cite{kosmos2} , Gemini~\cite{gemini}. Among them, MiniGPT4 performs worst due to the restricted leverage of visual information compared to text. Despite demonstrating notable cross-modal reasoning and visual grounding capabilities, both Gemini and Kosmos-2 face prominent challenges in the global context analysis. We also select several SOTA backbones to explore and validate the effectiveness of SatireDecoder, including LLaVA~\cite{liu2024llava}, LLaVA-NeXT~\cite{liu2024llavanext}, Qwen-VL~\cite{Qwen-VL}, and Qwen2.5-VL~\cite{Qwen2.5-VL}. The inference optimization with uncertainty analysis is conducted with the temperature hyperparameters from 0.2 to 1.0.


\textbf{Comparison of Baselines.} Primarily, we perform the automatic evaluation to conduct a fair comparison among the baseline models and our 
 proposed method, using the natural language generation (NLG) metrics, including BLEU~\cite{bleu}, ROUGE-L~\cite{lin-2004-rouge}, METEOR~\cite{meteor}, BERTScore~\cite{bertscore}, and an average score of the four NLG metrics.

As the result shown in Table~\ref{tab-1}, the baseline models MiniGPT4, GPT4, Kosmos-2, and Gemini exhibit disappointing performance. The average scores (normalized between 0 and 1) of automatic evaluation of baseline models are below 0.34, while the MLLM backbones equipped with SatireDecoder surpass the base models by approximately 4\%. Furthermore, as the results indicate, our proposed SatireDecoder demonstrates a clear superiority in the NLG metrics, which assess the extent of n-gram matching, semantic correspondence, variation in vocabulary, syntactic patterns, and the completeness of essential information between the generated and reference texts, providing a comprehensive assessment of the satire comprehension quality.

\textbf{User Study.} To evaluate whether the satire comprehension more accurately aligns with human intentions, we sample 100 images from YesBut, conducting a user study of the generated content by three users with a binary ``yes'' or ``no'' judgment. The evaluation is based on four criteria: correctness of the satire interpretation, appropriateness of the interpretation length, visual completeness interpretation, and faithfulness to the visual objects. Each criterion is evaluated by three users with a binary ``yes'' or ``no'' judgment.
As shown in Table~\ref{tab-1}, the baseline models encounter difficulties in accurately interpreting the satirical meanings within images, resulting in interpretations that lack precision in capturing the visual nuances. In contrast, our approach significantly surpasses the baseline models and backbones in the correctness of satire comprehension and the integrity and faithfulness of visual information representation by approximately 10\% to 40\%, effectively mitigating hallucinations in satire comprehension. By incorporating SatireDecoder during inference, LLaVA achieves significant improvements in correctness, completeness, and faithfulness, enhancing the above metrics by approximately 37\%, 20\%, and 33\%, respectively, with minimal computational overhead.

\begin{figure*}[htbp] 
    \centering
    \includegraphics[width=\textwidth]{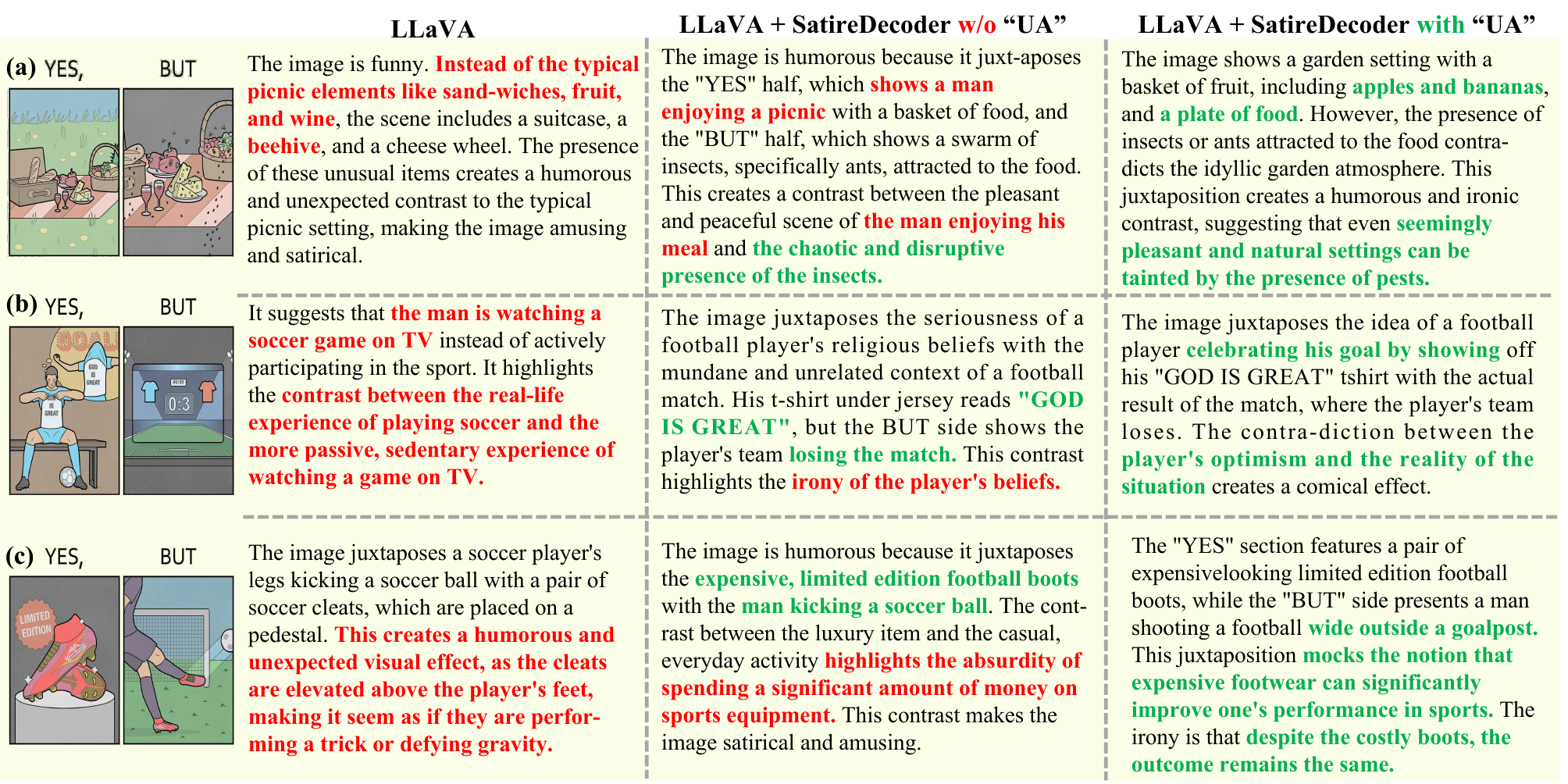} 
    \caption{Visualization of the ablation study. ``UA'' represents Uncertainty Analysis. The \textbf{red text} indicates the hallucinations and misinterpretations of satirical meaning. The \textbf{green text} highlights the objects and the correct satirical meanings newly captured after the application of SatireDecoder and uncertainty analysis.
}
    \label{fig-visualize}
\end{figure*}

\textbf{Ablation Study.}
To validate the effectiveness of uncertainty analysis in SatireDecoder, we conduct an ablation study. As shown in Table~\ref{tab-2}, the user study results indicate that uncertainty analysis significantly enhanced the performance of backbones in terms of the correctness of the satire comprehension, the visual completeness of the generated text, and the faithfulness to the visual objects. Specifically, the improvements are approximately 6\% to 20\% for correctness, about 10\% to 16\% for visual completeness, and roughly 12\% to 18\% for faithfulness. 

Furthermore, to validate the effectiveness on mitigating hallucinations, we employ the CHAIR metric~\cite{chairmetric} to measure the object hallucinations arising in the satire comprehension. CHAIR metric calculates the proportion of generated words that accurately correspond to the local objects in an image, as determined by the ground truth sentences and object segmentations. CHAIR is extended into two variants: $CHAIR\_i$, which represents the fraction of object instances that are hallucinated, and $CHAIR\_s$, which stands for the fraction of sentences that include a hallucinated object. Let \(H_o\) and \(Num_o\) denote hallucinated objects and all objects mentioned. And let \(H_s\) and \(Num_s\) stand for sentences with hallucinated objects and all sentences. CHAIR metrics are computed as:

\begin{equation}
\text{CHAIR}\_i = \frac{|H_o|}{|Num_o|}
\end{equation}

\begin{equation}
\text{CHAIR}\_s = \frac{|H_s|}{|Num_s|}
\end{equation}

As shown in Table~\ref{tab-2}, uncertainty analysis plays an important role in reducing the CHAIR metrics. The experimental results reveal that by restricting the inclusion of extraneous semantic entities in generated text during the inference phase, uncertainty analysis significantly reduces both object-level and sentence-level hallucinations, thereby enhancing the correctness of satirical image comprehension.

\begin{table}[ht]
\setlength{\tabcolsep}{6pt}
\centering

\setlength{\tabcolsep}{3pt} 
\begin{tabular}{@{}lcccc@{}}
\toprule
\textbf{Model} & \textbf{Corr.}~$\uparrow$ & \textbf{Len.}~$\uparrow$ & \textbf{Comp.}~$\uparrow$ & \textbf{Faith.}~$\uparrow$ \\ 
\midrule
LLaVA+$\clubsuit$           & \textbf{62.33} & \textbf{21.33} & \textbf{42.67} & \textbf{59.67} \\
LLaVA+$\clubsuit$  (w/o LE) & 50.33 & 20.33 & 37.67 & 38.33 \\
LLaVA+$\clubsuit$  (w/o GS) & 47.67 & 18.67 & 34.00 & 41.33 \\
LLaVA+$\clubsuit$  (w/o DA) & 54.00 & 19.67 & 38.33 & 42.67 \\
\bottomrule
\end{tabular}

\caption{Ablation study for multi-agent system in visual cascaded decoupling. ``LE'', ``GS'', and ``DA'' stand for Local Entities Agent, Global Semantics Agent, and Discrepancy Analysis Agent, respectively. ``Corr.'', ``Len.'', ``Comp.'', and ``Faith.'' stand for the four criteria in the user study.}

\label{tab-3}

\end{table}
To validate the contribution of each component in the multi-agent system, we also conduct an ablation study. As shown in Table~\ref{tab-3}, each part of our multi-agent collaboration is beneficial for visual cascaded decoupling and satire comprehension.

\textbf{Visualization.} To visually demonstrate the effectiveness of our proposed method, we compare the results of SatireDecoder and baseline models, as shown in Figure~\ref{fig-visualize}. 
GPT4 and LLaVA either overlook or fabricate crucial local entities within images, resulting in misinterpretations of the deep semantics in satirical images. However, LLaVA equipped with SatireDecoder shows some improvement in detecting local entities and details, and further integration with uncertainty analysis strategy significantly enhances the satire comprehension capabilities, capturing more fine-grained details missed by baseline models. As shown in Figure~\ref{fig-visualize}~(a), LLaVA equipped SatireDecoder (without uncertainty analysis) exhibits object-level hallucination ``the man enjoying his meal'', which does not exist in the image. Under the effect of uncertainty analysis, the above hallucination has been killed. Moreover, as illustrated in Figure~\ref{fig-visualize}~(c), only LLaVA equipped with SatireDecoder and uncertainty analysis successfully captures the spatial position relationship of the soccer ball with the goalpost, indicating no goal, a crucial nuance that other models fail to recognize. More visual comparisons are presented in the Appendix submitted with Supplementary Materials.
The improved ability allows MLLMs to better understand the relationship between local entities and the global context, leading to a more accurate comprehension of satirical meaning.

\section{Conclusion}
We propose a training-free approach, SatireDecoder, to enhance MLLMs' perception and comprehension of satirical images, addressing the challenging multimodal satire comprehension task. Experiments demonstrate the effectiveness of multi-agent-based visual cascaded decoupling in boosting the perception of visual information, as well as the utility of CoT-based uncertainty analysis in mitigating hallucinations, making SatireDecoder a powerful and cost-effective approach for satirical image comprehension.

\appendix

\section{Acknowledgments}
This project was funded by the National Natural Science Foundation of China (82090052).

\bibliography{aaai2026}

\end{document}